\documentclass[runningheads]{llncs}
\usepackage{graphicx}
\usepackage{comment}
\usepackage{amsmath,amssymb} % define this before the line numbering.
\usepackage{color}
\usepackage{multirow}
\usepackage{cite}
\usepackage{subfig}
\usepackage{epsfig}
\usepackage{algorithm}
\usepackage{algorithmic}

\usepackage[font=footnotesize,skip=0pt]{caption}

\begin{document}
%\renewcommand\thelinenumber{\color[rgb]{0.2,0.5,0.8}\normalfont\sffamily\scriptsize\arabic{linenumber}\color[rgb]{0,0,0}}
%\renewcommand\makeLineNumber {\hss\thelinenumber\ \hspace{6mm} \rlap{\hskip\textwidth\ \hspace{6.5mm}\thelinenumber}}
% \linenumbers
\pagestyle{headings}
\mainmatter
\def\ECCVSubNumber{509}  % Insert your submission number here

\title{Solving Long-tailed Recognition with Deep Realistic Taxonomic Classifier}

\begin{comment}
\titlerunning{ECCV-20 submission ID \ECCVSubNumber} 
\authorrunning{ECCV-20 submission ID \ECCVSubNumber} 
\author{Anonymous ECCV submission}
\institute{Paper ID \ECCVSubNumber}
\end{comment}
%******************

%\begin{comment}
\titlerunning{Deep Realistic Taxonomic Classifier} 
\author{Tz-Ying Wu \and Pedro Morgado \and Pei Wang \and Chih-Hui Ho \and Nuno Vasconcelos}

%\author{First Author\inst{1}\orcidID{0000-1111-2222-3333} \and
%Second Author\inst{2,3}\orcidID{1111-2222-3333-4444} \and
%Third Author\inst{3}\orcidID{2222--3333-4444-5555}}
\authorrunning{T.Y. Wu et al.}
\institute{University of California, San Diego \\
\email{\{tzw001, pmaravil, pew062, chh279, nvasconcelos\}@ucsd.edu}}
%\end{comment}
%******************
\maketitle

\begin{abstract}
Long-tail recognition tackles the natural non-uniformly distributed data in real-world scenarios. While modern classifiers perform well on populated classes, its performance degrades significantly on tail classes. Humans, however, are less affected by this since, when confronted with uncertain examples, they simply opt to provide coarser predictions. Motivated by this, a \textit{deep realistic taxonomic classifier} (Deep-RTC) is proposed as a new solution to the long-tail problem, combining realism with hierarchical predictions. The model has the option to reject classifying samples at different levels of the taxonomy, once it cannot guarantee the desired performance. Deep-RTC is implemented with a stochastic tree sampling during training to simulate all possible classification conditions at finer or coarser levels and a rejection mechanism at inference time. Experiments on the long-tailed version of four datasets, CIFAR100, AWA2, Imagenet, and iNaturalist, demonstrate that the proposed approach preserves more information on all classes with different popularity levels. Deep-RTC also outperforms the state-of-the-art methods in longtailed recognition, hierarchical classification, and learning with rejection literature using the proposed \textit{correctly predicted bits} (CPB) metric.
\end{abstract}

\section{Introduction}
\label{sec.intro}
Recent advances in computer vision can be attributed to large datasets~\cite{imagenet} and deep convolutional neural networks (CNN) \cite{alexnet,vggnet,resnet}. While these models have achieved great success on balanced datasets, with approximately the same number of images per class, real world data tends to be highly imbalanced, with a very long-tailed class distribution. In this case, classes are frequently split into many-shot, medium-shot and few-shot, based on the number of examples~\cite{oltr}. Since deep CNNs tend to overfit in the small data regime, they frequently underperform for medium and few-shot classes.  
Popular attempts to overcome this limitation include data resampling \cite{Han2005,abs-1710-05381,Drummond2003C4, Chawla2002}, cost-sensitive losses \cite{cbloss}, knowledge transfer from high to low population classes~\cite{oltr,NIPS2017_7278}, normalization~\cite{iclr2020}, or margin-based methods~\cite{cao2019learning}. All these approaches seek to improve the classification performance of the standard softmax CNN architecture. 

\begin{figure}[t!]
    \centering
    \includegraphics[width=0.65\linewidth]{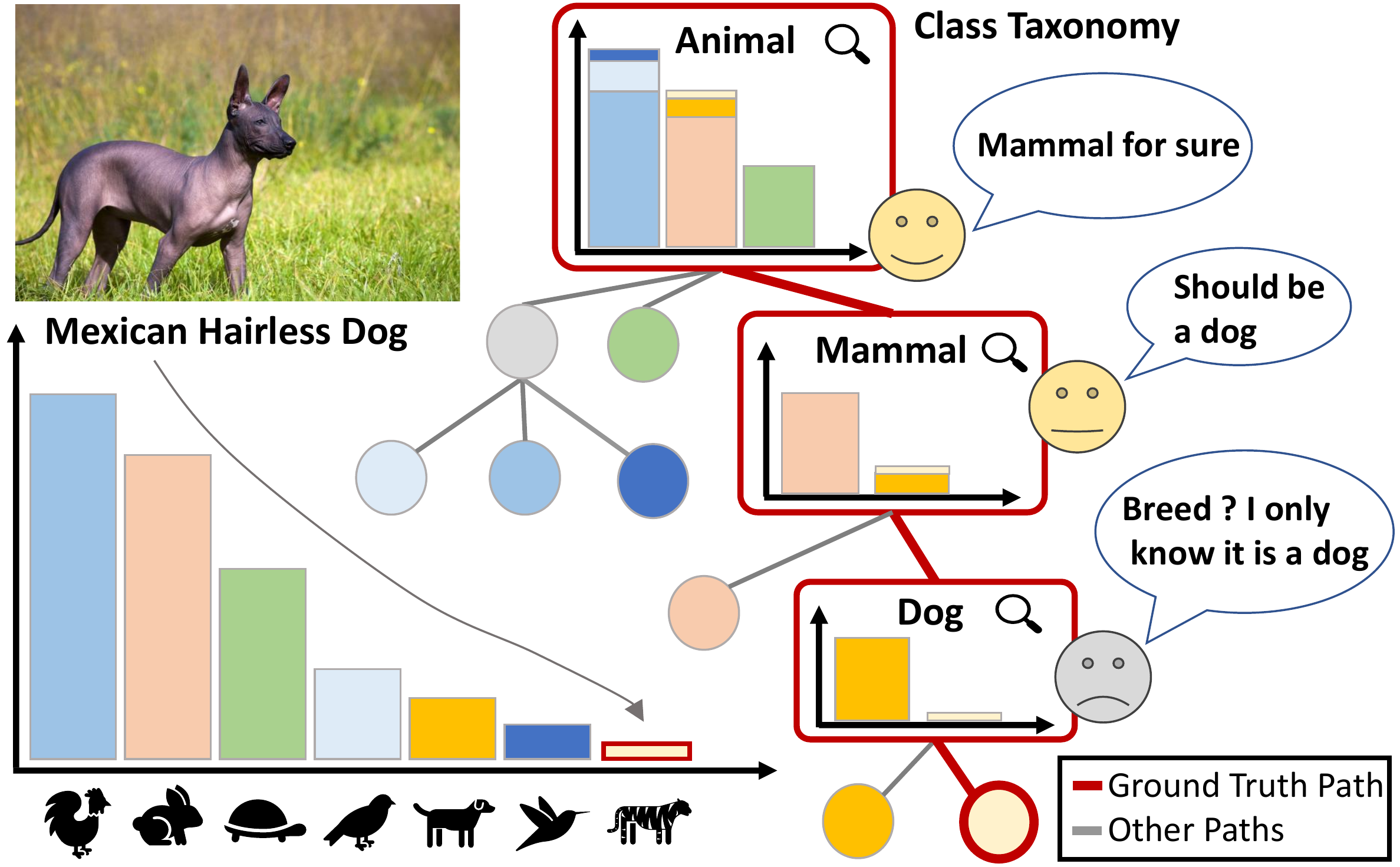}
    \caption{Real-world datasets have class imbalance and long tails (left). Humans deal with these problems by combining class taxonomies and self-awareness (right). When faced with rare objects, like a ``Mexican Hairless Dog", they push the decision to a coarser taxonomic level, e.g., simply recognizing a ``Dog", of which they feel confident. This is denoted as {\it realistic taxonomic classification\/} to guarantee that all samples are processed with a high level of confidence. }
    \label{fig.1}
\end{figure}

There is, however, little evidence that this architecture is optimally suited to deal with long-tailed recognition. For example, humans do not use this model. Rather than striving for discrimination between all objects in the world, they adopt class {\it taxonomies\/}~\cite{objectcate,impact,adaptive,effects,familiarity}, where classes are organized hierarchically at different levels of granularity, e.g.~ranging from coarse domains to fine-grained `species' or `breeds,' as shown in Figure~\ref{fig.1}. Classification with taxonomies is broadly denoted as {\it hierarchical\/}. The standard softmax, also known as the {\it flat,\/} classifier is a hierarchical classifier of a single taxonomic level. The use of deeper taxonomies has been shown advantageous for classification by allowing feature sharing~\cite{hd-cnn, taxonomy-regularized, b-cnn, vt-cnn, network-of-experts, Kim18} and information transfer across classes~\cite{hier-object-recog, large-scale-cate, 42854,share-appearance, hier-prior}. While most previous works on either flat or hierarchical classification attempt to classify all images at the leaves of the taxonomic tree, independently of how difficult this is, the introduction of a taxonomy enables alternate strategies.

In this work, we explore a strategy inspired by human cognition and suited for long-tailed recognition. When humans feel insufficiently trained to answer a question at a certain level of granularity, they simply provide an answer to a coarser level, for which they feel {\it confident.\/} For example, most people do not recognize the animal of Figure~\ref{fig.1} as a ``Mexican Hairless Dog". Instead, they change the problem from classifying dog breeds into classifying mammals and simply say it is a ``Dog". Hence, a long-tailed recognition strategy more consistent with human cognition is to adopt hierarchical classification and allow decisions at {\it intermediate\/} tree levels, to achieve two goals: 1) classify all examples with high confidence, and 2) classify each example as deep in the tree as possible without violating the first goal. Since examples from low-shot classes are harder to classify confidently than those of popular classes, they tend to be classified at earlier tree levels. This can be seen as a soft version of {\it realistic classification\/}~\cite{learn-with-reject,real-pred} where a classifier refuses to process examples of low-classification confidence and is denoted {\it realistic taxonomic classification\/} (RTC). The taxonomic extension enables multiple ``exit levels'' for the classification, at different taxonomic levels. 

RTC recognizes that, while classification at the leaves uncovers full label information, {\it partial\/} label information can still be recovered when this is not feasible, by performing the classification at intermediate taxonomic stages. The goal is then to maximize the {\it average\/} information recovered per sample, favoring correct decisions of intermediate level over incorrect decisions at the leaves. We introduce a new measure of classifier performance, denoted {\it correctly predicted bits\/} (CPB), to capture this average information and propose it as a new performance measure for long-tailed recognition. Rather than simply optimizing classification accuracy at the leaves, high CPB scores require learning algorithms that produce calibrated estimates of class probabilities at {\it all\/} tree levels. This is critical to enable accurate determination of when examples should leave the tree. For long-tailed recognition, where different images can be classified at different taxonomic levels, this calibration is particularly challenging.

We address this problem with two new contributions to the training of deep CNNs for RTC. The first is a new regularization procedure based on {\it stochastic tree sampling\/} (STS), which allows the consideration of all possible cuts of the taxonomic tree during training. RTC is then trained with a procedure similar to dropout~\cite{Dropout}, which considers the CNNs consistent with all these cuts. The second contribution addresses the challenge that RTC requires a {\it dynamic\/} CNN, capable of generating predictions at different taxonomic levels for each input example. This is addressed with a novel {\it dynamic predictor synthesis\/} procedure inspired by parameter inheritance, a regularization strategy commonly used in hierarchical classification~\cite{share-appearance, hier-prior}. To the best of our knowledge, these contributions enable the first implementation of RTC with deep CNNs and dynamic predictors. This is denoted as {\it Deep-RTC\/}, which achieves leaf classification accuracy comparable to state of the art long-tail recognition methods, but recovers much more average label information per sample. 

Overall, the paper makes three contributions. 1) we propose RTC as a new solution to the long-tailed problem. 2) the {\it Deep-RTC\/} architecture, which implements a combination of stochastic taxonomic regularization and dynamic taxonomic prediction, for implementation of RTC with deep CNNs.  3) an alternative setup for the evaluation of long-tailed recognition, based on CPB scores, that accounts for the amount of information in class predictions.

\section{Related Work}\label{sec.rw}
This work is related to several previously explored topics.  

{\bf Long-Tailed Recognition:} 
Several strategies have been proposed to address class unbalance in recognition. One possibility is to perform data resampling~\cite{5128907}, by undersampling head and oversampling tail classes~\cite{Han2005,abs-1710-05381,Drummond2003C4, Chawla2002}. Sample synthesis~\cite{4633969,Zou_2018_ECCV} has also been proposed to increase the population of tail classes. Unlike Deep-RTC, these methods do not seek improved classification architectures for long-tailed recognition. An alternative is to transfer knowledge from head to tail classes. Inspired by meta-learning~\cite{NIPS2016_6408,Wang2016LearningTL}, these methods learn how to leverage knowledge from head classes to improve the generalization of tail classes~\cite{NIPS2017_7278}. \cite{oltr} introduces memory features that encapsulate knowledge from head classes and uses an attention mechanism to discriminate between head and tail classes. This has some similarity with Deep-RTC, which also transfers knowledge from head to tail classes, but does so by leveraging hierarchical relations between them. Long-tailed recognition has also been addressed with cost-sensitive losses, which assign different weights to different samples. A typical approach is to weight classes by their frequency~\cite{huang2016lmle, Dhruv18} or treat tail classes as hard examples~\cite{Qi17}. \cite{cbloss} proposed a class balanced loss that can be directly applied to a softmax layer and focal loss~\cite{FocalLoss}. These approaches can underperform for very low-frequency classes. \cite{cao2019learning} addressed this problem by enforcing large margins for few-shot classes, where the margin is inversely proportional to the number of class samples. While effective losses for long-tailed recognition are a goal of this work, we seek losses for calibration of taxonomic classifiers, which cost-sensitive losses do not address. Finally, inspired by the correlation between the weight norm of a class and its number of samples, \cite{iclr2020} proposed to adjust the former after classifier training. All these approaches use the flat softmax classifier architecture and do not address the design of RTC.

{\bf Hierarchical Classification:} Hierarchical classification has received substantial attention in computer vision. For example, sharing information across classes has been used for object recognition on large and unbalanced datasets~\cite{hier-object-recog,large-scale-cate, 42854}, and defining a common hierarchical semantic space across classes has been explored for zero-shot learning~\cite{hier-zero-shot,Akata_2015_CVPR}. Some of the ideas used in this work, e.g. parameter inheritance, are from this literature~\cite{share-appearance,hier-prior,hedge-your-bets}. However, most of them precede deep learning and cannot be directly applied to modern CNNs. More recently, the ideas of sharing parameters or features hierarchically have inspired the design of CNN architectures~\cite{hd-cnn,taxonomy-regularized,b-cnn, vt-cnn,network-of-experts,Kim18}. Some of these do not support class taxonomies, e.g. learning hierarchical feature structures for flat classification~\cite{hier-zero-shot,Kim18}. Others are only applicable to a somewhat rigid two-level hierarchy~\cite{network-of-experts,hd-cnn}. Closer to this work are architectures that complement a flat classifier with convolutional branches that regularize its features to enforce hierarchical structure \cite{taxonomy-regularized,b-cnn,vt-cnn}. These branches can be based on hierarchies of feature pooling operators~\cite{taxonomy-regularized}, or classification layers~\cite{b-cnn,vt-cnn} supervised with labels from intermediate taxonomic levels. However, the use of additional layers makes the comparison to flat classifier unfair, which would undermine an important goal of the paper: to investigate the benefit of hierarchical (over flat) classification for long-tailed recognition.
Hence, we avoid hierarchical architectures that add parameters to the backbone network. These methods also fail to address a central challenge of RTC, namely the need for simultaneous optimization with respect to many label sets, associated with the different levels of the class taxonomy. This requires a dynamic network, whose architecture can change on-the-fly to enable 1) the use of different label sets to classify different samples, and 2) optimization with respect to many label sets.

{\bf Learning with Rejection}
The idea of learning with rejection dates back to at least \cite{chow1}. Subsequent works derive theoretical results on the error-rejection trade-off~\cite{chow2, noisefree-sc}, and explore alternative rejection criteria that avoid computation of class posterior probabilities \cite{svm-rej1, boost-rej, learn-with-reject}. Since the introduction of deep learning has made the estimation of the posterior distribution central to classification, most recent rejection functions consist of thresholding posteriors or derived quantities, such as the posterior entropy~\cite{sel-cls-deep, selectivenet, real-pred}. Alternative rejection methods have also been proposed, including the use of relative distances between samples~\cite{NIPS2018_7798}, Monte-Carlo dropout \cite{mc-dropout}, or classification model with a routing or rejection network~\cite{real-pred, selectivenet,NIPS2019_8556}. We adopt the simple threshold based rejection rule of~\cite{sel-cls-deep, selectivenet, real-pred} in our implementation of RTC. However, rejection is applied to each level of a hierarchical classifier, instead of once for a flat classifier. 
This resembles the hedge your bets strategy of~\cite{hedge-your-bets, isvc19}, in that it aims to maximize the average label information recovered per sample. However, while~\cite{hedge-your-bets, isvc19} accumulate the class probabilities of a flat classifier, our Deep-RTC addresses the calibration of probabilities {\it throughout\/} the tree. 
Our experiments show that this significantly outperforms the accumulation of flat classifier probabilities.
\cite{isvc19} further calibrates class probabilities before rejection, but calibration is only conducted a posteriori (at test time). Instead, we propose STS for training hierarchical classifiers whose predictions are inherently calibrated at all taxonomic levels.

\section{Long-tailed recognition and RTC}\label{sec:motivation}
This section motivates the need for RTC as a solution to long-tailed recognition.

{\bf Long-tailed Recognition} 
Existing approaches formulate long-tailed recognition as flat classification, solved by some variant of the softmax classifier. This combines a feature extractor $h(\mathbf{x};\mathbf{\Phi})\in \mathbb{R}^k$, implemented by a CNN of parameters $\mathbf{\Phi}$, and a softmax regression layer composed by a linear transformation $\mathbf{W}$ and a softmax function $\sigma(\cdot)$ 
\begin{equation}
f(\mathbf{x};\mathbf{W},\mathbf{\Phi}) =\sigma(z(\mathbf{x};\mathbf{W},\mathbf{\Phi}))
\quad \quad \quad 
z(\mathbf{x};\mathbf{W},\mathbf{\Phi}) 
= \mathbf{W}^{T} h(\mathbf{x};\mathbf{\Phi}).
    \label{eq.pflat}
\end{equation}
These networks are trained to minimize classification errors. Since samples are limited for mid and low-shot classes, performance can be weak. Long-tailed recognition approaches address the problem with example resampling, cost-sensitive losses, parameter sharing across classes, or post-processing. These strategies are not free of drawbacks. For example, cost-sensitive or resampling methods face a ``whack-a-mole" dilemma, where performance improvements in low-shot classes (e.g. by giving them more weight) imply decreased performance in more populated ones (less weight). They are also very different from the recognition strategies of human cognition, which relies extensively on class taxonomies.

Many cognitive science studies have attempted to determine taxonomic levels at which humans categorize objects \cite{objectcate,impact,adaptive,effects,familiarity}. This has shown that most object classes have a default level, which is used by most humans to label the object (e.g. ``dog'' or ``cat''). However, this so-called basic level is known to vary from person to person, depending on the person's training, also known as {\it expertise\/}, on the object class \cite{objectcate,effects,familiarity}. For example, a dog owner naturally refers to his/her pet as a ``labrador'' instead of as ``dog.'' This suggests that even humans are not great long-tail recognizers. Unless they are experts (i.e. have been extensively trained in a class), they instead perform the classification at a higher taxonomic level. 
From a machine learning point of view, this is sensible in two ways. First, by moving up the taxonomic tree, it is always possible to find a node with sufficient training examples for accurate classification. Second, while not providing full label information for all examples, this is likely to produce a higher average label information per sample than the all-or-nothing strategy of the flat classifier~\cite{hedge-your-bets, isvc19}. In summary, when faced with low-shot classes, humans {\it trade-off classification granularity for class popularity,\/} choosing a classification level where their training has enough examples to guarantee accurate recognition. This does not mean that they cannot do fine-grained recognition, only that this is reserved for classes where they are experts. For example, because all humans are extensively trained on face recognition, they excel in this very fine-grained task. These observations  motivate the RTC approach to long-tailed recognition.

\begin{figure}[t!]
    \centering
    \includegraphics[width=0.95\linewidth]{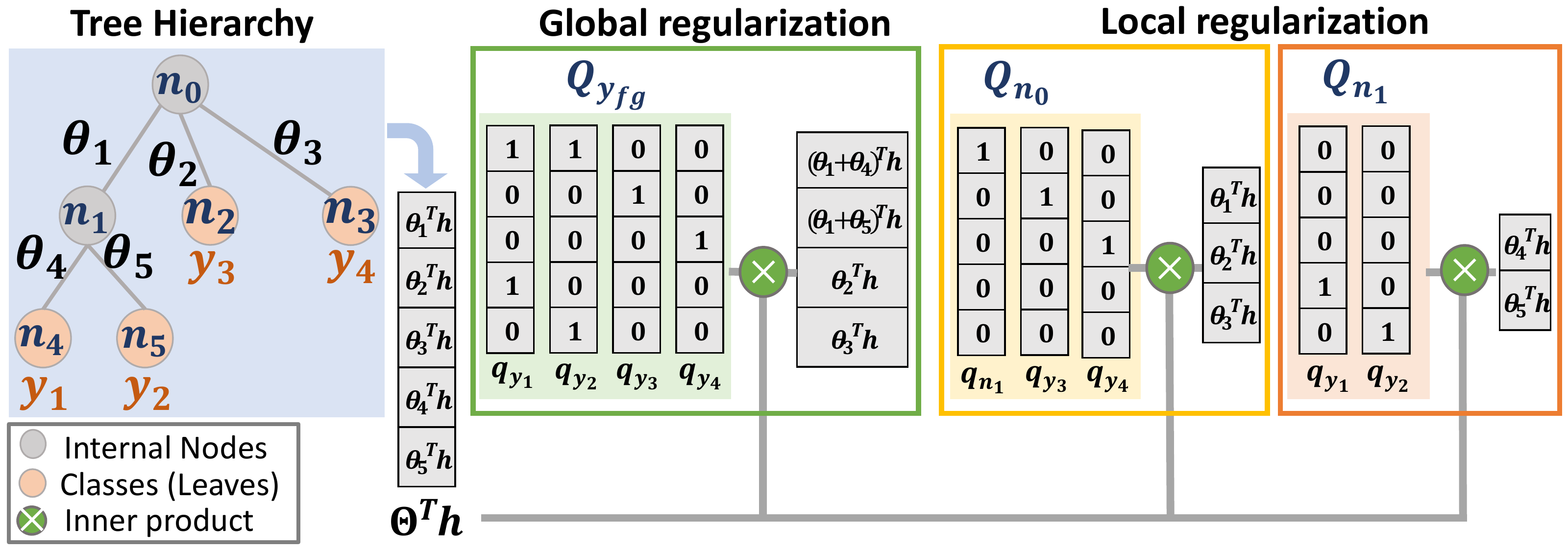}
    \caption{Parameter sharing based on the tree hierarchy are implemented through the codeword matrices $Q$. The training is regularized globally from the stochastically selected label set and locally from the node-conditional consistency loss.}
    \label{fig.hier}
\end{figure}

{\bf Realistic Taxonomic Classification}
A taxonomic classifier maps images $\mathbf{x}\!\in\!\mathcal{X}$ into a set of $C$ classes $y \in \mathcal{Y}\!\in\!\{1, \ldots, C\}$, organized into a taxonomic structure where classes are recursively grouped into parent nodes according to a tree-type hierarchy $\mathcal{T}$. It is defined by a set of classification nodes $\mathcal{N}\!=\!\{n_1, \cdots,n_N\}$ and a set of taxonomic relations $\mathcal{A}\!=\!\{\mathcal{A}(n_1), \cdots,\mathcal{A}(n_N)\}$, where $\mathcal{A}(n)$ is the set of ancestor nodes of $n$. The finest-grained classification decisions admitted by the taxonomy occur at the leaves. We denote this set of fine-grained classes $\mathcal{Y}_{fg}\!=\!\mbox{Leaves}(\mathcal{T})$. Figure~\ref{fig.hier} gives an example for a classification problem with $|\mathcal{Y}_{fg}|\!=\!4, |\mathcal{N}|\!=\!5$, $\mathcal{A}(n_4) = \mathcal{A}(n_5) = \{n_1\}$ and $\mathcal{A}(n_i) = \emptyset, i \in \{1, 2, 3\}$. Classes $y_1, y_2$ belong to parent class $n_1$ and the root $n_0$ is a dummy node containing all classes. Note that we use $n$ to represent nodes and $y$ to represent leaf labels. In RTC, {\it different samples can be classified at different hierarchy levels.\/} For example, a sample of class $y_2$ can be rejected at the root, classified at node $n_1$, or classified into one of the leaf classes. These options assign successively finer-grained labels to the sample. Samples rejected at the root can belong to any of the four classes, while those classified at node $n_1$ belong to classes $y_1$ or $y_2$. Classification at the leaves assigns the sample to a single class. 
Hence, RTC can predict any sub-class in the taxonomy $\mathcal{T}$. Given a training set  $\mathcal{D}=\{(\mathbf{x}_i, y_i)\}_{i=1}^{M}$ of images and class labels, and a class taxonomy $\mathcal{T}$, the {\it goal\/} is to learn a pair of classifier $f({\bf x})$ and rejection function $g({\bf x})$ that work together to assign each input image $\mathbf{x}$ to the finest grained class $\hat{y}$ possible, while guaranteeing certain confidence in this assignment.

\begin{figure*}[t!]
    \centering
    \includegraphics[width=\linewidth]{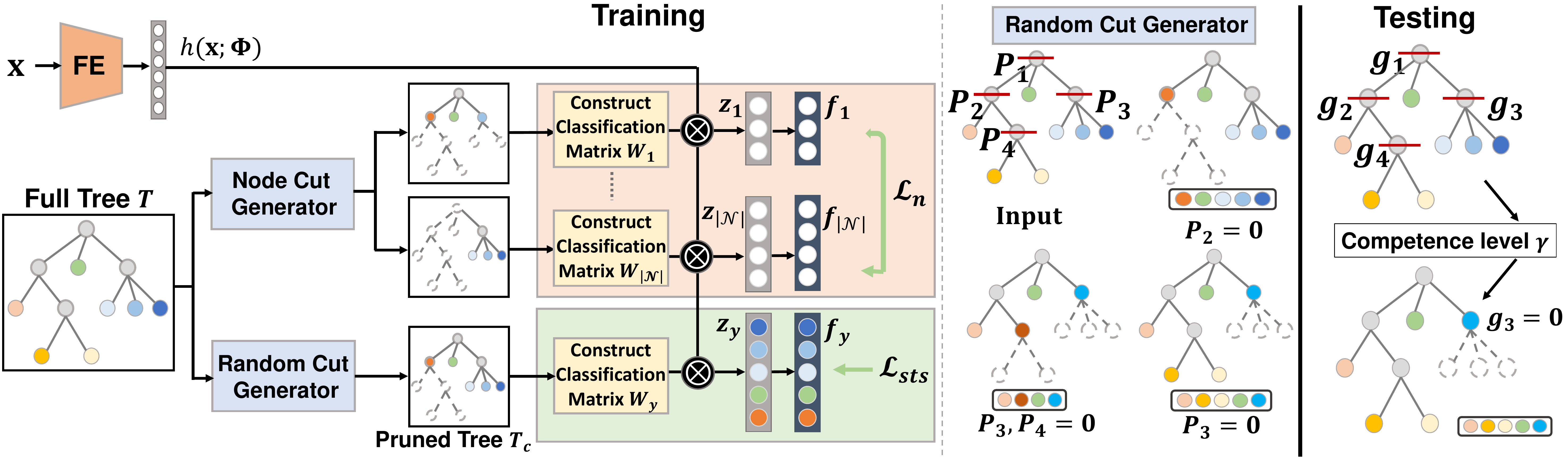}
    \caption{Left: Deep-RTC is composed of a feature extractor, a node cut generator producing $\mathcal{Y}_n\!=\!\mathcal{C}(n)$ for all internal nodes and a random cut generator producing a potential label sets $\mathcal{Y}_c$ from $\mathcal{T}_c$. Classification matrix $\mathbf{W}_{\mathcal{Y}_c}$ is constructed for each label set and loss of (\ref{eq:loss}) is imposed. 
    Right: Rejecting samples at certain level during inference time. 
    }
    \label{fig.model}
\end{figure*}

The depth at which the class prediction $\hat{y}$ is made depends on the sample difficulty and the {\it competence-level\/} $\gamma$ of the classification. This is a lower bound for the confidence with which $\mathbf{x}$ can be classified. A confidence score $s(f({\bf x}))$ is defined for $f({\bf x})$, which is declared competent (at the $\gamma$ level) for classification of $\mathbf{x}$, if $s(f({\bf x})) \geq \gamma$. RTC has competence level $\gamma$ if all its intermediate node decisions have this competence level. While this may be impossible to guarantee for classification with the leaf label set $\mathcal{Y}_{fg}$, it can always be guaranteed by rejecting samples at intermediate nodes of the hierarchy, i.e. defining 

\begin{equation}
    g_v(\mathbf{x};\gamma) = 1_{[s(f_v(\mathbf{x}))\geq \gamma]}
\end{equation}
per classification node $v$, where $1_{[.]}$ is the Kroneker delta. 
This prunes the hierarchy $\mathcal{T}$ {\it dynamically\/} per sample ${\bf x}$, producing a customized cut $\mathcal{T}_p$ for which the hierarchical classifier is competent at a competence level $\gamma$.
This pruning is illustrated on the right of Figure~\ref{fig.model}.
Samples that are hard to classify, e.g. from few-shot classes, induce low confidence scores and are rejected earlier in the hierarchy. Samples that match the classifier expertise, e.g. from highly populated classes, progress until the leaves. This is a generalization of flat realistic classifiers~\cite{real-pred}, which simply accept or reject samples. RTC mimics human behavior in that, while $\mathbf{x}$ may not be classified at the finest-grained level, confident predictions can usually be made at intermediate or coarse levels. The competence level $\gamma$ offers a guarantee for the quality of these decisions. Since larger values of $\gamma$ require decisions of higher confidence, they encourage sample classification early in the hierarchy, avoiding the harder decisions that are more error-prone. The trade-off between accuracy and fine-grained labeling is controlled by adjusting $\gamma$. The confidence score $s(\cdot)$ can be implemented in various ways~\cite{real-pred,selectivenet,NIPS2019_8556}. While RTC is compatible with any of these, we adopt the popular maximum posterior probability criterion, i.e. ${s(f({\bf x})) = \max_i f^i ({\bf x})}$, where $f^i(\cdot)$ is the $i^{th}$ entry of $f(.)$. 
In our experience, the calibration of the node predictors $f_v ({\bf x})$ is more important than the particular implementation of the confidence score function.

\section{Taxonomic probability calibration}
In this section, we introduce the architecture of Deep-RTC. 

{\bf Taxonomic calibration}
Since RTC requires decisions at all levels of the taxonomic tree, samples can be classified into any potential label set $\mathcal{Y}$ containing leaf nodes of any cut of $\mathcal{T}$.
For example, the taxonomy of Figure~\ref{fig.hier} admits two label sets, namely, $\mathcal{Y}_{fg}=\{y_1, y_2, y_3, y_4\}$ containing all classes and $\mathcal{Y}=\{n_1, y_3, y_4\}$ obtained by pruning the children of node $n_1$. For long-tailed recognition, where different images can be classified at very different taxonomic levels, it is important to calibrate the posterior probability distributions of {\it all\/} these label sets. We address this problem by  optimizing the ensemble of {\it all} classifiers implementable with the hierarchy, i.e., minimize the loss
\begin{align}
    \textstyle\mathcal{L}_{ens} = \frac{1}{|\Omega|} \sum_{\mathcal{Y} \in \Omega } L_{\mathcal{Y}}~,
    \label{eq:loss_ens}
\end{align}
where $\Omega$ is the set of all target label sets $\mathcal{Y}$ that can be derived from $\mathcal{T}$ by pruning the tree and $L_{\mathcal{Y}}$ is a loss function associated with label set $\mathcal{Y}$.
While feasible for small taxonomies, this approach does not scale with taxonomy size, since the set $\Omega$ increases exponentially with $|\mathcal{T}|$. 
Instead, we introduce a mechanism, inspired by dropout~\cite{Dropout}, for {\it stochastic tree sampling (STS)\/} during training.
At each training iteration, a random cut $\mathcal{T}_c$ of the taxonomy $\mathcal{T}$ is sampled, and the predictor $f_{\mathcal{Y}_c}(\mathbf{x};\mathbf{W}_{\mathcal{Y}_c}, \mathbf{\Phi})$ associated with the corresponding label set $\mathcal{Y}_c$ is optimized.
For this, random cuts are generated by sampling a Bernoulli random variable $P_v \sim Bernoulli(p)$ for each internal node $v$ with a given dropout rate $p$. The subtree rooted at $v$ is pruned if $P_v=0$. Examples of these taxonomy cuts are shown in Figure~\ref{fig.model}. The predictor $f_{\mathcal{Y}_c}$ of~(\ref{eq.pflat}) consistent with the target label set $\mathcal{Y}_c$ associated with the cut $\mathcal{T}_c$ is then synthesized, and the loss computed as 
\begin{align}
    \textstyle\mathcal{L}_{sts} = \frac{1}{M}\sum_{i=1}^{M}L_{\mathcal{Y}_c}(\mathbf{x}_i, y_i).
    \label{eq:sploss}
\end{align}
By considering different cuts at different iterations, the learning algorithm forces the hierarchical classifier to produce well calibrated decisions for all label sets.

{\bf Parameter sharing} The procedure above requires {\it on-the-fly\/} synthesis of predictors $f_{\mathcal{Y}_c}$ for all possible label sets $\mathcal{Y}_c$ that can be derived from taxonomy $\mathcal{T}$. This implies a {\it dynamic\/} CNN architecture, where~\eqref{eq.pflat} changes with the sample $\mathbf{x}$. Deep-RTC is one such architecture, inspired by the fact that, for long-tailed recognition, the predictors $f_{\mathcal{Y}_c}$ should share parameters, so as to enable information transfer
from head to tail classes~\cite{NIPS2017_7278,oltr}. This is implemented with a combination of two parameter sharing mechanisms. First, the backbone feature extractor $h(\mathbf{x};\mathbf{\Phi})$ is shared across all label sets.
Since this enables the implementation of Deep-RTC with a single network and no additional parameters, it is also critical for fair comparisons with the flat classifier. More complex hierarchical network architectures \cite{taxonomy-regularized, b-cnn, vt-cnn} would compromise these comparisons and are not investigated. Second, the predictor of~\eqref{eq.pflat} should reflect the hierarchical structure of each label set $\mathcal{Y}_c$. A popular implementation of this constraint, denoted {\it parameter inheritance (PI),\/} reuses parameters of ancestors nodes $\mathcal{A}(n)$ in the predictor of node $n$. The column vector ${\bf w}_n$ of ${\bf W}_{\mathcal{Y}}$ is then defined as
\begin{align}
    \textstyle {\bf w}_n = {\bf\theta}_n + \sum_{p\in\mathcal{A}(n)} {\bf\theta}_p~,
    \quad\forall n\in\mathcal{Y}
    \label{eq:pred_node}
\end{align}
where ${\bf\theta}_n$ are non-hierarchical node parameters. 
This compositional structure has two advantages. First, it leverages the parameters of parent nodes (more training data) to regularize the parameters of their low-level descendants (less training data).  Second, the parameter vector ${\bf \theta}_n$ of node $n$ only needs to model the residuals between $n$ and its parent, in order to be discriminative of its siblings. In summary, low-level decisions are simultaneously simplified and robustified.

{\bf Dynamic predictor synthesis}
Deep-RTC is a novel architecture to enable the {\it dynamic\/} synthesis of predictors $f_{\mathcal{Y}_c}$ that comply with~\eqref{eq:pred_node}. This is achieved by introducing a codeword vector ${\bf q}_n\in \{0,1\}^{|\mathcal{N}|}$ per node $n$, containing binary flags that identify the ancestors $\mathcal{A}(n)$ of $n$
\begin{equation}
    \mathbf{q}_n(v) = 1_{[v\in \mathcal{A}(n)\cup\{n\}]}.
\end{equation}
For example, in the taxonomy of Figure~\ref{fig.hier}, ${\bf q}_{n_1}\!=\!(1, 0, 0, 0, 0)$ since $\mathcal{A}(n_1)\!=\!\varnothing$, and ${\bf q}_{n_4}\!=\!(1, 0, 0, 1, 0)$ since $\mathcal{A}(n_4)\!=\!\{n_1\}$. Codeword ${\bf q}_n$ encodes which nodes of $\mathcal{T}$ contribute to the prediction of node $n$ under the PI strategy, thus providing a recipe for composing predictors for any label set $\mathcal{Y}$. A matrix of node-specific parameters $\mathbf{\Theta}=[\theta_1, \ldots, \theta_{|\mathcal{N}|}]$ where $\theta_n\in\mathbb{R}^k$ for all $n\in\mathcal{N}$ is then introduced, and 
${\bf w}_n$ can be reformulated as
\begin{align}
    \textstyle{\bf w}_n = \mathbf{\Theta} {\bf q}_n.
    \label{eq:pred}
\end{align}
The codeword vectors of all nodes $n\in\mathcal{Y}$ are then written into the columns of a codeword matrix ${\bf Q}_{\mathcal{Y}}\in \{0,1\}^{|\mathcal{N}|\times|\mathcal{Y}|}$, to define a predictor 
as in \eqref{eq.pflat},
\begin{equation}
f_{\mathcal{Y}}(\mathbf{x};\mathbf{\Theta}, \mathbf{\Phi}) =\sigma(z_{\mathcal{Y}}(\mathbf{x};\mathbf{\Theta}, \mathbf{\Phi})) 
\quad \quad \quad 
z_{\mathcal{Y}}(\mathbf{x};\mathbf{\Theta}, \mathbf{\Phi}) = \mathbf{W}_{\mathcal{Y}}^T h({\bf x}; {\bf \Phi}),
\label{eq.zY}
\end{equation}
where $\mathbf{W}_{\mathcal{Y}}=\mathbf{\Theta}{\bf Q}_{\mathcal{Y}}$.
This enables the classification of sample $\mathbf{x}$ with respect to {\it any\/} label set $\mathcal{Y}_c$ by simply making $\mathbf{Q}_{\mathcal{Y}}$ 
a dynamic matrix $\mathbf{Q}_{\mathcal{Y}}(\mathbf{x}) =\mathbf{Q}_{\mathcal{Y}_c}$,
as illustrated in Figure~\ref{fig.model}. 

{\bf Loss function} Deep-RTC is trained with a cross-entropy loss
\begin{align}
    L_{\mathcal{Y}}(\mathbf{x}_i, y_i) = -\mathbf{y}_{i}^T\log f_{\mathcal{Y}}(\mathbf{x};\mathbf{\Theta}, \mathbf{\Phi})~,
    \label{eq:xent}
\end{align}
where $\mathbf{y}_i$ is the one-hot encoding of $y_i\in\mathcal{Y}$. When this is used in~\eqref{eq:sploss}, the CNN is globally optimized with respect to the label set 
$\mathcal{Y}_c$ associated with taxonomic cut $\mathcal{T}_c$. The regularization of the many classifiers associated with different cuts of $\mathcal{T}$ is a {\it global\/} regularization, guaranteeing that all classifiers are well calibrated. 
Beyond this, it is also possible to calibrate the internal node-conditional  decisions. Given that a sample $\mathbf{x}$ has been assigned to node $n$, the node-conditional decisions are {\it local\/} and determine  which of the children $\mathcal{C}(n)$ the sample should be assigned to. They consider only the target label set $\mathcal{Y}_n\!=\!\mathcal{C}(n)$ defined by the children of $n$. 
For these label sets, all nodes $v\in\mathcal{C}(n)$ share the same ancestor set $\mathcal{A}_v$ and thus the second term of (\ref{eq:pred_node}). Hence, after softmax normalization, (\ref{eq:pred_node}) is equivalent to ${\bf w}_v = {\bf\theta}_v$ and the node-conditional classifier $f_n(\cdot)$ reduces to
\begin{align}
    &f_n(\mathbf{x};\mathbf{\Theta},\mathbf{\Phi})=
    \sigma(\mathbf{Q}_n^{T} \, \mathbf{\Theta}^T\, h(\mathbf{x};\mathbf{\Phi})),
\end{align}
where, as illustrated in Figure~\ref{fig.hier}, the codeword matrix $\mathbf{Q}_n$ contains zeros for all ancestor nodes.
Internal node decisions can thus be calibrated by noting that sample ${\bf x}_i$ provides supervision for all node-conditional classifiers in its ground-truth ancestor path $\mathcal{A}(y_i)$. This allows the definition of a node-conditional consistency loss per node $n$ of the form
\begin{align}
    \textstyle
    \mathcal{L}_n = 
    \frac{1}{M} \sum_{i=1}^M 
    \frac{1}{|\mathcal{A}(y_i)|}\sum_{n\in \mathcal{A}(y_i)}
    L_{\mathcal{Y}_n}(\mathbf{x}_i, y_{n,i})
    \label{eq:Ln}
\end{align}
where $L_{\mathcal{Y}_n}$ is the loss of \eqref{eq:xent} for the label set $\mathcal{Y}_n$ and $y_{n,i}$ the label of ${\bf x}_i$ for the decision at node $n$. Deep-RTC is trained by minimizing a combination of these local node-conditional consistency losses and the global ensemble loss of~\eqref{eq:sploss}
\begin{align}
    \mathcal{L}_{cls} =  \mathcal{L}_n + \lambda \mathcal{L}_{sts},
    \label{eq:loss}
\end{align}
where $\lambda$ weights the contribution of the two terms.

{\bf Performance Evaluation} Due to the universal adoption of the flat classifier, previous  long-tailed recognition works equate performance to recognition accuracy. Under the taxonomic setting, this is identical to measuring leaf node accuracy $\mathbb{E}\{1_{[\hat{y}_i = y_i]}\}$ and fails to reward trade-offs  
between classification granularity and accuracy. In the example of Figure~\ref{fig.1}, it only rewards the ``Mexican Hairless Dog" label, making no distinction between the labels ``Dog" or ``Tarantula," which are both considered errors. A taxonomic alternative is to rely on  hierarchical accuracy $\mathbb{E}\{1_{[\hat{y}_i\in\mathcal{A}(y_i)]}\}$ \cite{hedge-your-bets}. This has the limitation of rewarding ``Dog" and ``Mexican Hairless Dog" equally, i.e. does not encourage finer-grained decisions. In this work, we propose that a better performance measure should capture the amount of class label information captured by the classification. While a correct classification at the leaves captures all the information, a rejection at an intermediate node can only capture partial information. To measure this, we propose to use the number of {\it correctly predicted bits\/} (CPB) by the classifier, under the assumption that each class at the leaves of the taxonomy contributes one bit of information. This is defined as 

\begin{equation}
    \textstyle
    \mbox{CPB} = \frac{1}{M} \sum_{i=1}^{M} 
    \mathrm{1}_{[\hat{y}_i\in \mathcal{A}(y_i)]}
    \left( 1 - \frac{\left|\mbox{Leaves}(\mathcal{T}_{\hat{y_i}})\right|}{\left|\mbox{Leaves}(\mathcal{T})\right|} \right)\label{eq.cpb}
\end{equation}
where $\mathcal{T}_{\hat{y_i}}$ is the sub-tree rooted at $\hat{y_i}$. This assigns a score of $1$ to correct classification at the leaves, and smaller scores to correct classification at higher tree levels. Note that any correct prediction of intermediate level is preferred to an incorrect prediction at the leaves, but scores less than a correct prediction of finer-grain. Finally, for flat classifiers, CPB is equal to classification accuracy.

\section{Experiments}
\label{sec.exp}
This section presents the long-tailed recognition performance of Deep-RTC.

\subsection{Experimental Setup}
\textbf{Datasets}. We consider 4 datasets. 
\textbf{CIFAR100-LT}\cite{cbloss}
is a long-tailed version of \cite{cifar100} with ``imbalance factor" $0.01$ (i.e. most populated class 100$\times$ larger than rarest class). \textbf{AWA2-LT} is a long-tailed version, curated by ourselves, of \cite{awa2}. It contains $30\,475$ images from $50$ animal classes and hierarchical relations  extracted from WordNet~\cite{wordnet}, leading to a 7-level imbalanced tree. The training set has an imbalance factor of $0.01$, the testing set is balanced. \textbf{ImageNet-LT}\cite{oltr} is a long-tailed version of \cite{imagenet}, with 1000 classes of more than 5 and less than 1280 images per class, and a balanced test set.
\textbf{iNaturalist (2018)} \cite{inat, inat2018} is a large-scale dataset of $8\,142$ classes with the class imbalance factor of $0.001$, and a balanced test set. While the full iNaturalist dataset is used for comparisons to previous work, a more manageable subset, iNaturalist-sub,  containing $55\,929$ images for training and $8\,142$ for testing, is used for ablation studies. Please refer to supplementary material for more details.

\noindent\textbf{Data partitions for long-tail evaluation}
The evaluation protocol of~\cite{oltr} is adopted by splitting the classes into many-shot, medium-shot, and few-shot. The  splitting rule of \cite{oltr} is used on iNaturalist. On CIFAR100-LT and AWA2-LT, the top and bottom $1/3$ populated classes belong to many-shot and few-shot respectively, and the remaining to medium-shot.

\noindent\textbf{Backbone architectures}
CIFAR100-LT and iNaturalist use the setup of \cite{cbloss}, where ResNet32 \cite{resnet} and ImageNet pre-trained ResNet50 are used respectively. For ImageNet-LT, ResNet10 is chosen as in \cite{oltr}. For AWA2-LT, we use ResNet18.

\noindent\textbf{Competence level} Unless otherwise noted, the value of $\gamma$ is cross-validated, i.e. the value of best performance on the validation set is applied to the test set.
\subsection{Ablations}
We started by evaluating how the different components of Deep-RTC - parameter inheritance (PI) regularization of~(\ref{eq:pred_node}), node-consistency loss (NCL) of (\ref{eq:Ln}), and stochastic tree sampling (STS) of (\ref{eq:xent}) - affect 
%long-tailed recognition
the performance.
Two baselines were used in this experiment. The first is a flat classifier, implemented with the standard softmax architecture, and trained to optimize classification accuracy. This is a representative baseline for the architectures used in the long-tailed recognition literature.  The second is a hierarchical classifier 
derived from this flat classifier, by recursively adding class probabilities as dictated by the class taxonomy. This is denoted as the {\it recursive hierarchical classifier\/} (RHC). We refer to this computation of probabilities as {\it bottom-up\/} (BU) inference. 
This is opposite to the {\it top-down\/} (TD) inference used by most hierarchical approaches, where probabilities are sequentially computed from the root (top) to leaves (bottom) of the tree. The performance of the different classifiers was measured in multiple ways. CPB is the metric of~(\ref{eq.cpb}). Leaf acc. is the classification accuracy at the leaves of the taxonomy. For a flat classifier, this is the standard performance measure. For a hierarchical classifier, it is the accuracy when intermediate rejections are not allowed. Hier acc. is the accuracy of a classifier that supports rejections, measured at the point where the decision is taken. In the example of Figure~\ref{fig.1} a decision of ``dog" is considered accurate under this metric. Finally, depth is the average depth at which images are rejected, normalized by tree depth (e.g. $1$ when no intermediate rejections are allowed).

\begin{figure*}[t]
\begin{minipage}[t]{\linewidth}
\begin{minipage}[h]{0.6\textwidth}
\captionof{table}{Ablations on iNaturalist-sub.}
 \centering
\resizebox{\linewidth}{!}{
 \setlength{\tabcolsep}{2.2pt}
 \begin{tabular}{lccccc }
 \hline
 Method & leaf acc. & depth & hier. acc. & CPB & inference \\\hline\hline
 Flat classifier & .163 & 1 & .163 & .163 & - \\
 RHC & .163 & .58 & .754 & .537  & BU \\ 
 PI+STS &  .174 & .46 & .913 & .601  & TD \\
 PI+NCL   &  .185 & .48 & .904 & .563  & TD\\
 PI+STS+NCL (Deep-RTC) & .181 & .50 & .899 & .619  & TD\\ 
 \hline
\end{tabular}\label{tab.ablation}
}
\vspace{1pt}
\captionof{table}{Comparisons to hierarchical classifiers.}
\resizebox{\linewidth}{!}{
\begin{tabular}{lcccc}
\hline
Method & CIFAR100-LT & AWA2-LT & ImageNet-LT & inference\\ \hline \hline
CNN-RNN~\cite{cnn-rnn} & .379 & .882 & .514 & TD \\
B-CNN~\cite{b-cnn} & .366 & .805 & .511 & TD/BU \\
HND~\cite{hier-novel-detect} & .374 &  - & - & TD \\
NofE~\cite{network-of-experts} & .373 & .770 & .463 & BU \\ \hline \hline
Deep-RTC & {\bf.397} & {\bf .894} & {\bf .529} & TD \\ \hline
\end{tabular}\label{tab.hiercls}
}
\end{minipage}
\begin{minipage}[h]{0.39\textwidth}
    \centering
    \includegraphics[width=\linewidth]{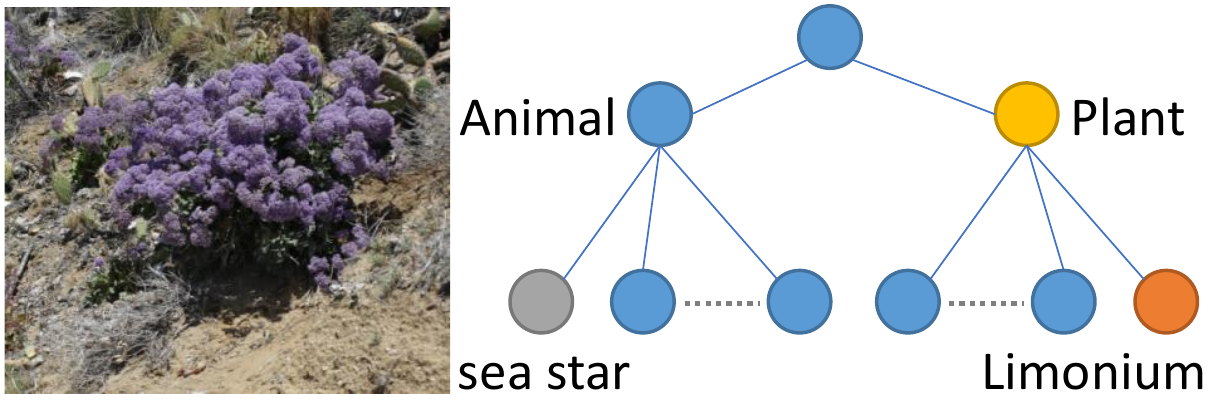} \\
    \includegraphics[width=\linewidth]{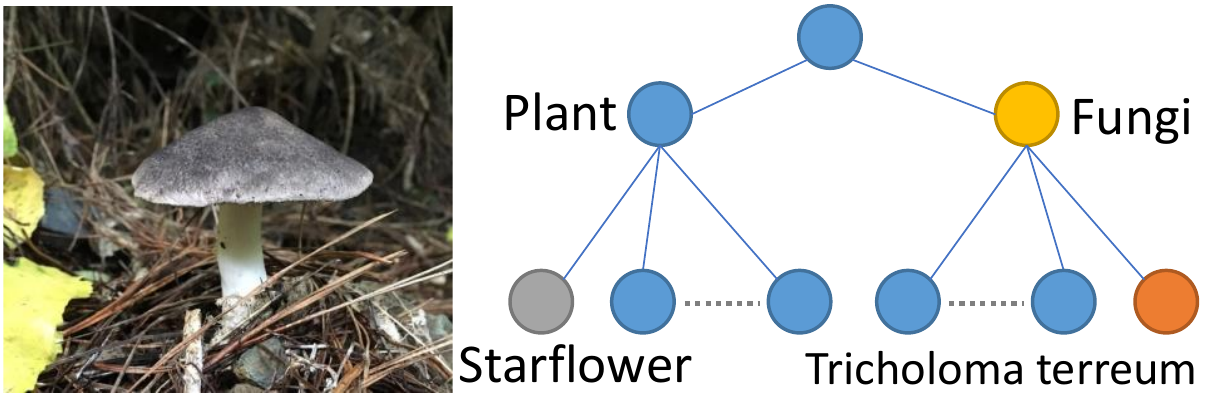}
    \captionof{figure}{Prediction of Deep-RTC (yellow) and flat classifer (gray) on two iNaturalist-sub images (orange: ground truth).}
    \label{fig.example}
\end{minipage}
\end{minipage}
\end{figure*}

\noindent{\bf CPB Performance:} Table~\ref{tab.ablation} shows that the flat classifier has very poor CPB performance
%. This is because prediction at the leaves with large class imbalance requires the classifier to make decisions on lots of classes where it is poorly trained. 
because prediction at leaves requires the classifier to make decisions on tail classes where it is poorly trained. 
The result is a very large number of errors, i.e. images for which no label information is preserved. RHC, its bottom-up hierarchical extension, is a much better solution to long-tailed recognition. While most images are not classified at the leaves, 
%hierarchical accuracy increases dramatically, and so does CPB.
both hierarchical accuracy and CPB increases dramatically.
Nevertheless, RHC has weaker CPB performance than the combination of the PI architecture of Figure~\ref{fig.hier} with either STS or NCL. Among these, the global regularization of STS is more effective than the local regularization of $L_n$. However, by combining two regularizations,
%the two regularizations are complementary. When combined, 
they lead to the classifier (Deep-RTC) that preserves most information about the class label.

\noindent{\bf Performance measures:} The long-tailed recognition literature has focused on maximizing the accuracy of flat classifiers. Table~\ref{tab.ablation} shows some limitations of this approach. First, all classifiers have very poor performance under this metric, with leaf acc. between $16\%$ and $18\%$. Furthermore, as shown in Figure~\ref{fig.example}, the labels can be totally uninformative of the object class. In the example, the flat classifier assigns the label of ``sea star" (``star flower") to the image of the plant (mushroom) shown on the top (at the bottom). We are aware of {\it no\/} application that would find such labels useful. Second, all classifiers perform dramatically better in terms of hier. acc. For the practitioner, this means that they are accurate classifiers. Not expert enough to always carry the decision to the bottom of the tree, but reliable in their decisions. 
In the same example, Deep-RTC instead correctly assigns the images to the broader classes of ``Plant" (top) and ``Fungi" (bottom).  
Furthermore, Deep-RTC classifies $90\%$ of the images correctly at this level! This could make it useful for many applications. For example, it could be used to automatically route the images to experts in these particular classes, for further labeling. Third, among TD classifiers, Deep-RTC pushes decisions furthest down the tree (e.g. $4\%$ deeper than PI+STS). This makes it a better {\it expert\/} on iNaturalist-sub than its two variants, a fact captured by the proposed CPB measure. Given all this, we believe that CPB optimality is much more meaningful than leaf acc. as a performance measure for long-tailed recognition.

\begin{table}[t!]
\begin{minipage}[t]{\textwidth}
\begin{minipage}[t]{0.64\textwidth}
\setlength{\tabcolsep}{5.5pt}
\centering
\caption{Results on iNaturalist. Classes are discussed with popularity classes (many, medium and few- shot).}\label{tab.norej_inat}
\resizebox{\linewidth}{!}{
\begin{tabular}{lc|ccc|c}
\hline
Method             & metric & Many & Medium & Few & All \\ \hline\hline
Softmax  & CPB & 0.76 & 0.67 & 0.62 & 0.66   \\
CBLoss~\cite{cbloss}  & & 0.61 & 0.62 & 0.61 & 0.61 \\
LDAM-SGD~\cite{cao2019learning} & & - & - & - & 0.65 \\
LDAM-DRW~\cite{cao2019learning} & & - & - & - & 0.68 \\
NCM~\cite{iclr2020} & & 0.61 & 0.64 & 0.63 & 0.63 \\
cRT~\cite{iclr2020} & & 0.73 & 0.69 & 0.66 & 0.68 \\
$\tau$-norm~\cite{iclr2020} & & 0.71 & 0.69 & 0.69 & 0.69   \\ \hline\hline
\multirow{4}{*}{Deep-RTC} & CPB & \textbf{0.84} & \textbf{0.79} & \textbf{0.75} & \textbf{0.78} \\
& hier. acc. & 0.92 & 0.91 & 0.89 & 0.90 \\
& leaf freq. & 0.71 & 0.56 & 0.48 & 0.54 \\
& leaf acc. & 0.76 & 0.67 & 0.60 & 0.64 \\\hline
\end{tabular}
}
\end{minipage}
%\end{table}
\hspace{0.02\textwidth}
\begin{minipage}[t]{0.34\textwidth}
%\begin{table}[t!]
\setlength{\tabcolsep}{5.5pt}
\centering
\caption{Results on ImageNet-LT.}\label{tab.norej_imgnet}
\resizebox{0.85\linewidth}{!}{
\begin{tabular}{lc}
\hline
Method             & CPB \\ \hline\hline
FSLwF~\cite{fslwf} & 0.28 \\ 
Focal Loss~\cite{FocalLoss} & 0.31 \\
Range Loss~\cite{rangeloss} & 0.31 \\ 
Lifted Loss~\cite{SongXJS15} & 0.31 \\
OLTR~\cite{oltr} & 0.36 \\
Softmax  & 0.35 \\
NCM~\cite{iclr2020} & 0.36 \\
cRT~\cite{iclr2020} & 0.42 \\
$\tau$-norm~\cite{iclr2020} & 0.41\\ \hline\hline
Deep-RTC & \textbf{0.53}   \\ \hline
\end{tabular}
}
\end{minipage}
\end{minipage}
\end{table}
\begin{table}[t]
\setlength{\tabcolsep}{5.5pt}
\centering
\caption{Comparisons to learning with rejection under different rejection rates (CPB).}\label{tab.rp}
%\caption{CPB on CIFAR100-LT and AWA2-LT for Deep-RTC and RP~\cite{real-pred}.}\label{tab.rp}
\resizebox{0.9\linewidth}{!}{%
\begin{tabular}{c|c|cccc|cccc}
    \hline
    & 
    & \multicolumn{4}{c|}{CIFAR100-LT} 
    & \multicolumn{4}{c}{AWA2-LT} \\
    Rej. Rate
    & Method
    & Many & Medium & Few & All 
    & Many & Medium & Few & All 
    \\ \hline \hline
    \multirow{2}{*}{$5\%$} & RP \cite{real-pred}
    & \textbf{.779} & \textbf{.722} & .306 & .404
    & \textbf{.977} & .963 & .887 & .914 \\
    & Deep-RTC
    & .773 & .719 & \textbf{.335} & \textbf{.416}
    & .975 & \textbf{.978} & \textbf{.907} & \textbf{.931}
    \\ \hline
    \multirow{2}{*}{$10\%$} & RP \cite{real-pred}
    & \textbf{.793} & \textbf{.734} & .315 & .416
    & \textbf{.980} & .966 & .900 & .924 \\
    & Deep-RTC
    & .789 & .7314 & \textbf{.344} & \textbf{.439}
    & .975 & \textbf{.984} & \textbf{.929} & \textbf{.947}
    \\ \hline
    \multirow{2}{*}{$20\%$} & RP \cite{real-pred}
    & .816 & .751 & .328 & .433
    & \textbf{.985} & .970 & .916 & .939 \\
    & Deep-RTC
    & \textbf{.833} & \textbf{.770} & \textbf{.393} & \textbf{.491}
    & .969 & \textbf{.975} & \textbf{.943} & \textbf{.954}
    \\ \hline
\end{tabular}
}
\end{table}

\subsection{Comparisons to hierarchical classifiers}
We next performed a comparison to prior works in hierarchical classification with CPB in Table~\ref{tab.hiercls}. 
These experiments show that prior methods have similar performance, without discernible advantage for TD or BU inference; however, they all underperform Deep-RTC. This is particularly interesting because these methods use networks more complex than Deep-RTC, adding branches (and parameters) to the backbone in order to regularize features according to the taxonomy. Deep-RTC simply implements a dynamic softmax classifier with the label encoding of Figure~\ref{fig.hier}. Instead, it leverages its dynamic ability and stochastic sampling to simultaneously optimize decisions for many tree cuts. The results suggest that this optimization over label sets is more important than shaping the network architecture according to the taxonomy. This is sensible since, under the Deep-RTC strategy, feature regularization is learned end-to-end, instead of hard-coded. Details of the compared methods are in the supplementary material.

\subsection{Comparisons to long-tail recognizers}\label{ssec.lt}
A comparison to the state of the art methods from the long-tailed recognition is presented in Tables~\ref{tab.norej_inat}-\ref{tab.norej_imgnet} for iNaturalist and ImageNet-LT respectively. More comparisons for other datasets are provided in the supplementary material.
In all cases, Deep-RTC predicts {\it more bits\/} correctly (i.e. higher CPB), which beats the state of the art flat classifier by $9\%$ on iNaturalist and $11\%$ on ImageNet-LT. For iNaturalist, we also discuss other metrics by class popularity, where leaf freq. represents the frequency that samples are classified to leaves. 
A comparison to the standard softmax classifier shows that prior long-tailed methods improve performance CPB on few-shot classes but {\it degrade\/} for popular classes.
Deep-RTC is the only method to consistently improve CPB performance for all levels of class popularity.
It is also noted that, unlike the state of the art flat classifier, Deep-RTC does not have to sacrifice leaf acc. for the many-shot classes in order to accommodate few-shot classes where its performance will not be great anyway. Instead, it exits early for about half of the images of the few-shot classes and guarantees highly accurate answers for all classes (around $90\%$ hier. acc.). This is similar to how humans treat the long-tail recognition problem.

\subsection{Comparisons to learning with rejection}
While the classifiers of the previous sections were allowed to reject examples at intermediate nodes, whenever feasible, they were not explicitly optimized for such rejection. Table~\ref{tab.rp} shows a comparison to a state-of-the-art flat realistic predictor (RP)~\cite{real-pred}, on CIFAR100-LT and AWA2-LT. In these comparisons, the percentage of rejected examples (rejection rate) is kept the same. The rejection rate of Deep-RTC is the percent of examples rejected at the root node. Deep-RTC achieves the best performance for all rejection rates on both datasets, because it has the option of soft-rejecting, i.e. letting examples propagate until some intermediate tree node. This is not possible for the flat RP, which always faces an all or nothing decision. In terms of class popularity, Deep-RTC always has higher CPB for few-shot classes, and frequently considerable gains. For many and medium-shot classes, the two methods have the comparable performance on CIFAR100-LT. On AWA2-LT, RP has an advantage for many and Deep-RTC for medium-shot classes. This shows that the gains of Deep-RTC are mostly due to its ability to push images of low-shot classes as far down the tree as possible without forcing decisions for which the classifier is poorly trained.

\section{Conclusion}\label{sec.con}
In this work, a \textit{realistic taxonomic classifier} (RTC) is proposed to address the long-tail recognition problem. Instead of seeking the finest-grained classification for each sample, we propose to classify each sample up to the level that the classifier is competent. Deep-RTC architecture is then introduced for implementing RTC with deep CNN and is able to
1) share knowledge between head and tail classes
2) align data hierarchy with model design in order to predict at all levels in the taxonomy, and 
3) guarantee high prediction performance by opting to provide coarser predictions when samples are too hard.
Extensive experiments validate the effectiveness of the proposed method on 4 long-tailed datasets using the proposed tree metric. This indicates that RTC is well suited for solving long-tail problem.
We believe this opens up a new direction for long-tailed literature.

\noindent \textbf{Acknowledgments} This work was partially funded by NSF awards IIS-1637941, IIS-1924937, and NVIDIA GPU donations.

\clearpage
%\bibliographystyle{splncs04}
%\bibliography{egbib}

\begin{thebibliography}{10}
\providecommand{\url}[1]{\texttt{#1}}
\providecommand{\urlprefix}{URL }
\providecommand{\doi}[1]{https://doi.org/#1}

\bibitem{inat2018}
inaturalist 2018 competition,
  \url{https://github.com/visipedia/inat\textunderscore comp}

\bibitem{network-of-experts}
Ahmed, K., Baig, M.H., Torresani, L.: Network of experts for large-scale image
  categorization. In: European Conference on Computer Vision (ECCV) (2016)

\bibitem{Akata_2015_CVPR}
Akata, Z., Reed, S., Walter, D., Lee, H., Schiele, B.: Evaluation of output
  embeddings for fine-grained image classification. In: IEEE Conference on
  Computer Vision and Pattern Recognition (CVPR) (2015)

\bibitem{familiarity}
Anaki, D., Bentin, S.: Familiarity effects on categorization levels of faces
  and objects. Cognition  (2009)

\bibitem{adaptive}
Anderson, J.: The adaptive nature of human categorization. Psychological Review
   (1991)

\bibitem{abs-1710-05381}
Buda, M., Maki, A., Mazurowski, M.: A systematic study of the class imbalance
  problem in convolutional neural networks. Neural Networks  \textbf{106} (10
  2017). \doi{10.1016/j.neunet.2018.07.011}

\bibitem{cao2019learning}
Cao, K., Wei, C., Gaidon, A., Arechiga, N., Ma, T.: Learning imbalanced
  datasets with label-distribution-aware margin loss. In: Advances in Neural
  Information Processing Systems (NIPS) (2019)

\bibitem{Chawla2002}
Chawla, N.V., Bowyer, K.W., Hall, L.O., Kegelmeyer, W.P.: Smote: Synthetic
  minority over-sampling technique. J. Artif. Int. Res.  \textbf{16}(1),
  321--357 (Jun 2002), \url{http://dl.acm.org/citation.cfm?id=1622407.1622416}

\bibitem{chow1}
Chow, C.K.: An optimum character recognition system using decision functions.
  IRE Transactions on Electronic Computers  \textbf{EC-6},  247 -- 254 (12
  1957)

\bibitem{chow2}
Chow, C.K.: On optimum recognition error and reject tradeoff. IEEE Transactions
  on Information Theory  \textbf{16},  41--46 (1 1970)

\bibitem{NIPS2019_8556}
Corbi\'{e}re, C., Thome, N., Bar-Hen, A., Cord, M., P\'{e}rez, P.: Addressing
  failure detection by learning model confidence. In: Advances in Neural
  Information Processing Systems (NIPS) (2019)

\bibitem{boost-rej}
Cortes, C., DeSalvo, G., Mohri, M.: Boosting with abstention. In: Advances in
  Neural Information Processing Systems (NIPS) (2016)

\bibitem{learn-with-reject}
Cortes, C., DeSalvo, G., Mohri, M.: Learning with rejection. In: International
  Conference on Algorithmic Learning Theory (ALT) (2016)

\bibitem{cbloss}
Cui, Y., Jia, M., Lin, T.Y., Song, Y., Belongie, S.: Class-balanced loss based
  on effective number of samples. In: IEEE Conference on Computer Vision and
  Pattern Recognition (CVPR) (2019)

\bibitem{isvc19}
Davis, J., Liang, T., Enouen, J., Ilin, R.: Hierarchical semantic labeling with
  adaptive confidence. In: International Symposium on Visual Computing (2019)

\bibitem{imagenet}
Deng, J., Dong, W., Socher, R., Li, L.J., Li, K., Fei-Fei, L.: Imagenet: A
  large-scale hierarchical image database. In: IEEE Conference on Computer
  Vision and Pattern Recognition (CVPR) (2009)

\bibitem{42854}
Deng, J., Ding, N., Jia, Y., Frome, A., Murphy, K., Bengio, S., Li, Y., Neven,
  H., Adam, H.: Large-scale object classification using label relation graphs.
  In: European Conference on Computer Vision (ECCV) (2014)

\bibitem{hedge-your-bets}
Deng, J., Krause, J., Berg, A.C., Fei-Fei, L.: Hedging your bets: Optimizing
  accuracy-specificity trade-offs in large scale visual recognition. In: IEEE
  Conference on Computer Vision and Pattern Recognition (CVPR) (2012)

\bibitem{Qi17}
Dong, Q., Gong, S., Zhu, X.: Class rectification hard mining for imbalanced
  deep learning. In: International Conference on Computer Vision (ICCV) (10
  2017)

\bibitem{Drummond2003C4}
Drummond, C., Holte, R.: C4.5, class imbalance, and cost sensitivity: Why
  under-sampling beats oversampling. Proceedings of the ICML'03 Workshop on
  Learning from Imbalanced Datasets  (01 2003)

\bibitem{noisefree-sc}
El-Yaniv, R., Wiener, Y.: On the foundations of noise-free selective
  classification. Journal of Machine Learning Research  \textbf{11},
  1605--1641 (5 2010)

\bibitem{svm-rej1}
Fumera, G., Roli, F.: Support vector machines with embedded reject option.
  Pattern recognition with support vector machines  \textbf{2388},  68–82 (7
  2002)

\bibitem{mc-dropout}
Gal, Y., Ghahramani, Z.: Dropout as a bayesian approximation: Representing
  model uncertainty in deep learning. In: International Conference on Machine
  Learning (ICML) (2016)

\bibitem{sel-cls-deep}
Geifman, Y., El-Yaniv, R.: Selective classification for deep neural networks.
  In: Advances in Neural Information Processing Systems (NIPS) (2017)

\bibitem{selectivenet}
Geifman, Y., El-Yaniv, R.: Selectivenet: A deep neural network with an
  integrated reject option. In: International Conference on Machine Learning
  (ICML) (2019)

\bibitem{fslwf}
Gidaris, S., Komodakis, N.: Dynamic few-shot visual learning without
  forgetting. In: IEEE Conference on Computer Vision and Pattern Recognition
  (CVPR) (06 2018)

\bibitem{taxonomy-regularized}
Goo, W., Kim, J., Kim, G., Hwang, S.J.: Taxonomy-regularized semantic deep
  convolutional neural networks. In: European Conference on Computer Vision
  (ECCV) (2016)

\bibitem{cnn-rnn}
Guo, Y., Liu, Y., Bakker, E.M., Guo, Y., Lew, M.S.: Cnn-rnn: a large-scale
  hierarchical image classification framework. Multimedia Tools and
  Applications  \textbf{77},  10251–10271 (2018)

\bibitem{4633969}
{Haibo He}, {Yang Bai}, {Garcia}, E.A., {Shutao Li}: Adasyn: Adaptive synthetic
  sampling approach for imbalanced learning. In: 2008 IEEE International Joint
  Conference on Neural Networks (IEEE World Congress on Computational
  Intelligence). pp. 1322--1328 (2008)

\bibitem{Han2005}
Han, H., Wang, W.Y., Mao, B.H.: Borderline-smote: A new over-sampling method in
  imbalanced data sets learning. Advances in Intelligent Computing
  \textbf{3644},  878--887 (09 2005)

\bibitem{5128907}
{He}, H., {Garcia}, E.A.: Learning from imbalanced data. IEEE Transactions on
  Knowledge and Data Engineering  \textbf{21}(9),  1263--1284 (Sep 2009).
  \doi{10.1109/TKDE.2008.239}

\bibitem{resnet}
He, K., Zhang, X., Ren, S., Sun, J.: Deep residual learning for image
  recognition. In: IEEE Conference on Computer Vision and Pattern Recognition
  (CVPR) (2016)

\bibitem{inat}
Horn, G.V., Aodha, O.M., Song, Y., Cui, Y., Sun, C., Shepard, A., abd
  Pietro~Perona, H.A., Belongie, S.: The inaturalist species classification and
  detection dataset. In: IEEE Conference on Computer Vision and Pattern
  Recognition (CVPR) (2018)

\bibitem{huang2016lmle}
Huang, C., Li, Y., Loy, C.C., Tang, X.: Learning deep representation for
  imbalanced classification. In: IEEE Conference on Computer Vision and Pattern
  Recognition (CVPR) (2016)

\bibitem{NIPS2018_7798}
Jiang, H., Kim, B., Guan, M., Gupta, M.: To trust or not to trust a classifier.
  In: Advances in Neural Information Processing Systems (NIPS). pp. 5541--5552
  (2018)

\bibitem{impact}
Johnson, K.: Impact of varying levels of expertise on decisions of category
  typicality. Memory \& Cognition  (2001)

\bibitem{effects}
Johnson, K., Mervis, C.: Effects of varying levels of expertise on the basic
  level of categorization. J Experimental Psychology: General  (1997)

\bibitem{iclr2020}
Kang, B., Xie, S., Rohrbach, M., Yan, Z., Gordo, A., Feng, J., Kalantidis, Y.:
  Decoupling representation and classifier for long-tailed recognition. In:
  International Conference on Learning Representations (ICLR) (2020)

\bibitem{Kim18}
Kim, H.J., Frahm, J.M.: Hierarchy of alternating specialists for scene
  recognition. In: European Conference on Computer Vision (ECCV) (2018)

\bibitem{cifar100}
Krizhevsky, A., Hinton, G.: Learning multiple layers of features from tiny
  images. Technical report, Citeseer  (2009)

\bibitem{awa2}
Krizhevsky, A., Hinton, G.: Zero-shot learning—a comprehensive evaluation of
  the good, the bad and the ugly. IEEE Transactions on Pattern Analysis and
  Machine Intelligence  \textbf{41},  2251 -- 2265 (9 2019)

\bibitem{alexnet}
Krizhevsky, A., Sutskever, I., Hinton, G.E.: Imagenet classification with deep
  convolutional neural networks. In: Advances in Neural Information Processing
  Systems (NIPS) (2012)

\bibitem{hier-novel-detect}
Lee, K., Lee, K., Min, K., Zhang, Y., Shin, J., Lee, H.: Hierarchical novelty
  detection for visual object recognition. In: IEEE Conference on Computer
  Vision and Pattern Recognition (CVPR) (2018)

\bibitem{FocalLoss}
Lin, T.Y., Goyal, P., Girshick, R., He, K., Dollar, P.: Focal loss for dense
  object detection. IEEE Transactions on Pattern Analysis and Machine
  Intelligence  \textbf{PP}, ~1--1 (07 2018). \doi{10.1109/TPAMI.2018.2858826}

\bibitem{vt-cnn}
Liu, Y., Dou, Y., Jin, R., Qiao, P.: Visual tree convolutional neural network
  in image classification. In: International Conference on Pattern Recognition
  (ICPR) (2018)

\bibitem{oltr}
Liu, Z., Miao, Z., Zhan, X., Wang, J., Gong, B., Yu, S.X.: Large-scale
  long-tailed recognition in an open world. In: IEEE Conference on Computer
  Vision and Pattern Recognition (CVPR) (2019)

\bibitem{Dhruv18}
Mahajan, D., Girshick, R., Ramanathan, V., He, K., Paluri, M., Li, Y.,
  Bharambe, A., van~der Maaten, L.: Exploring the limits of weakly supervised
  pretraining. In: European Conference on Computer Vision (ECCV) (2018)

\bibitem{hier-object-recog}
Marszałek, M., Schmid, C.: Semantic hierarchies for visual object recognition.
  In: IEEE Conference on Computer Vision and Pattern Recognition (CVPR) (2007)

\bibitem{wordnet}
Miller, G.A.: Wordnet: A lexical database for english. Communications of the
  ACM  \textbf{38},  39--41 (11 1995)

\bibitem{hier-zero-shot}
Morgado, P., Vasconcelos, N.: Semantically consistent regularization for
  zero-shot recognition. In: IEEE Conference on Computer Vision and Pattern
  Recognition (CVPR) (2017)

\bibitem{share-appearance}
Salakhutdinov, R., Torralba, A., Tenenbaum, J.: Learning to share visual
  appearance for multiclass object detection. In: IEEE Conference on Computer
  Vision and Pattern Recognition (CVPR) (2011)

\bibitem{hier-prior}
Shahbaba, B., Neal, R.M.: Improving classification when a class hierarchy is
  available using a hierarchy-based prior. Bayesian Analysis, 2(1):221–238
  (2007)

\bibitem{vggnet}
Simonyan, K., Zisserman, A.: Very deep convolutional networks for large-scale
  image recognition. CoRR  \textbf{abs/1409.1556} (2014)

\bibitem{SongXJS15}
Song, H.O., Xiang, Y., Jegelka, S., Savarese, S.: Deep metric learning via
  lifted structured feature embedding. In: IEEE Computer Vision and Pattern
  Recognition (CVPR) (2016)

\bibitem{Dropout}
Srivastava, N., Hinton, G., Krizhevsky, A., Sutskever, I., Salakhutdinov, R.:
  Dropout: A simple way to prevent neural networks from overfitting. Journal of
  Machine Learning Research  \textbf{15}(56),  1929--1958 (2014),
  \url{http://jmlr.org/papers/v15/srivastava14a.html}

\bibitem{objectcate}
Tanaka, J., Taylor, M.: Object categories and expertise: Is the basic level in
  the eye of the beholder. Cognitive Psychology  (1991),
  \url{https://doi.org/10.1016/0010-0285(91)90016-H}

\bibitem{real-pred}
Wang, P., Vasconcelos, N.: Towards realistic predictors. In: European
  Conference on Computer Vision (ECCV) (2018)

\bibitem{NIPS2016_6408}
Wang, Y.X., Hebert, M.: Learning from small sample sets by combining
  unsupervised meta-training with cnns. In: Advances in Neural Information
  Processing Systems (NIPS) (2016)

\bibitem{Wang2016LearningTL}
Wang, Y.X., Hebert, M.: Learning to learn: Model regression networks for easy
  small sample learning. In: European Conference on Computer Vision (ECCV)
  (2016)

\bibitem{NIPS2017_7278}
Wang, Y.X., Ramanan, D., Hebert, M.: Learning to model the tail. In: Advances
  in Neural Information Processing Systems (NIPS) (2017)

\bibitem{hd-cnn}
Yan, Z., Zhang, H., Piramuthu, R., Jagadeesh, V., DeCoste, D., Di, W., Yu, Y.:
  Hd-cnn: Hierarchical deep convolutional neural networks for large scale
  visual recognition. In: International Conference on Computer Vision (ICCV)
  (2015)

\bibitem{rangeloss}
Zhang, X., Fang, Z., Wen, Y., Li, Z., Qiao, Y.: Range loss for deep face
  recognition with long-tailed training data. In: International Conference on
  Computer Vision (ICCV) (2017)

\bibitem{large-scale-cate}
Zhao, B., Fei-Fei, L., Xing, E.P.: Large-scale category structure aware image
  categorization. In: Advances in Neural Information Processing Systems (NIPS)
  (2011)

\bibitem{b-cnn}
Zhu, X., Bain, M.: B-cnn: Branch convolutional neural network for hierarchical
  classification. CoRR  \textbf{abs/1709.09890} (2017)

\bibitem{Zou_2018_ECCV}
Zou, Y., Yu, Z., Vijaya~Kumar, B., Wang, J.: Unsupervised domain adaptation for
  semantic segmentation via class-balanced self-training. In: European
  Conference on Computer Vision (ECCV) (2018)

\end{thebibliography}

\end{document}